\documentclass[conference]{IEEEtran}
\usepackage{graphicx}
\usepackage{epstopdf}

\ifCLASSINFOpdf
\else
\fi

\begin{document}
\title{$k$-Modulus Method for Image Transformation}
\author{\IEEEauthorblockN{Firas A. Jassim}
\IEEEauthorblockA{Department of Management Information Systems\\
Irbid National University\\
Irbid 2600, Jordan\\
Email: firasajil@yahoo.com}
}
\maketitle

\begin{abstract}
%\boldmath
In this paper, we propose a new algorithm to make a novel spatial image transformation. The proposed approach aims to reduce the bit depth used for image storage. The basic technique for the proposed transformation is based of the modulus operator. The goal is to transform the whole image into multiples of predefined integer. The division of the whole image by that integer will guarantee that the new image surely less in size from the original image. The $k$-Modulus Method could not be used as a stand alone transform for image compression because of its high compression ratio. It could be used as a scheme embedded in other image processing fields especially compression. According to its high PSNR value, it could be amalgamated with other methods to facilitate the redundancy criterion.

\end{abstract}
\IEEEpeerreviewmaketitle

\section{Introduction}
% no \IEEEPARstart
Image transformations have received a significant consideration due to its importance in computer vision, computer graphics, and medical imaging \cite{WOLBERG1988Geo}. The transformed images are information about the dissimilarities between input image and the output image and trying to make these dissimilarities as closely as possible \cite{Kaczmarek_comparisonof}. Image transformations are the extremely significant process in image processing and image analysis. Image transforms are used in image enhancement, restoration, reconstruction, encoding and description. A spatial transformation is an image processing process that re-identifies the spatial relationship between pixels in an image. This procedure will make the manipulation of an image layout is easier concerning image size and shape. The mathematical concept of image transformation is a powerful procedure that is recruited in various fields in image processing disciplines and the most important one is image compression \cite{Salomon:2007:DCC}. The pixels pattern in natural images is not random, but has a noticeable statistical regularity. This regularity is clearly may be touched at the single pixel level or the block of pixels \cite{DBLP:conf/cvpr/MemisevicH07}. Redundancy in data is exemplary reachable through transforming the original data from one representation to another \cite{Acharya:2004:JSI:1027521}. The basic idea in image compression is that the new pixels are almost smaller than the original pixels. Alternatively, the decoder side enters the transformed pixels and reconstructs the original image by implementing the inverse transform procedure \cite{Salomon:2007:DCC}. The image transformation could be imagined as a mapping from one coordinates to another transformed coordinates.
It must be mentioned that, there are two types of image transformations according to the reversibility procedure. The first one is lossless transform and the other one is the lossy transform. In case of lossless transform, the original pixels are completely reversible. While in the lossy transforms, the genuine pixels could not be obtained because lossy transformations are fully irreversible \cite{Gonzalez:2001:DIP:559707}.
In principle, the idea behind image transformation is the representation of the pixels in the original image in terms of fewer bit length than the original bit stream. Transformation of an image usually decreases the entropy of the original image by eliminating the redundancies of the image pixel sequence \cite{Acharya:2004:JSI:1027521}.

The remainder of the paper is organized as follows. Section II discusses the spatial transformation for image processing. The proposed $k$-Modulus Method ($k$-MM) has been introduced by details in section III. Section IV presents the experimental results when applying the proposed $k$-Modulus Method to a variety of test images. The conclusion of this paper was presented in section V.

\section{Spatial Transformation Preliminaries}
Spatial transformations are important in many aspects of functional image analysis \cite{Friston:2006:Ashburner}. The main concept in spatial transformation is to find a mapping for each pixel lies in the original integer lattice to a new grid used to resample the input pixels \cite{WOLBERG1988Geo}. The relationship between the image and its corresponding mapping is depending on the structure of the images themselves \cite{DBLP:conf/cvpr/MemisevicH07}. Therefore, the selection of the suitable representation of image is only part of the solution for transforming the original image into a suitable mapping for ulterior processing \cite{Gonzalez:2001:DIP:559707}.
According to \cite{WOLBERG1988Geo}, the geometric transformation is completely determined by the spatial transformation because analytic mapping is bijective, i.e. one-to-one and onto. The principle motivation behind transformation could be summarized in two aspects. The firs aspect is that, transformation yielding more efficient representation of the original samples. On the other hand, the transformed pixels must demand fewer bits \cite{Rao:2007:DCT:1200193}. Spatial transformation mappings may take different forms and this is depends on the type of application \cite{WOLBERG1988Geo}.
The choice of the mapping model is essential to obtain reliable results; this choice must be considered seriously and assessed before any processing. As mentioned above, the transformation functions depend on the nature of the scene and on the acquisition system used. However, under some assumptions, and before applying a complicated model, it may be useful to reach the idealistic mapping solicited by consecutive less and less 'rigid' mapping models \cite{Chanussot:2009:DIP:Multivariate}.

The general mapping function can be given in the following form. Let $f$ be the original image and $g$ is the transformed image then there exist an operator $T$ such that:

\begin{equation}
T:f \longrightarrow g,\,\,\,\,\,\,\,\,\,\, f,g \in R^{N\times M}
\end{equation}
and
\begin{equation}
g(x,y)=T\left[f(x,y)\right]
\end{equation}
where $N$ and $M$ are the image dimensions. It must be mentioned that the operator $T$ is a one-to-one operator, i.e. for each pixel in $f$ there exist a corresponding one and only one pixel is $g$. Moreover, the one-to-one transformation could be shown in figure \ref{fig:fig}.

\begin{figure}[h]
	\centering
		\includegraphics[width=0.45\textwidth]{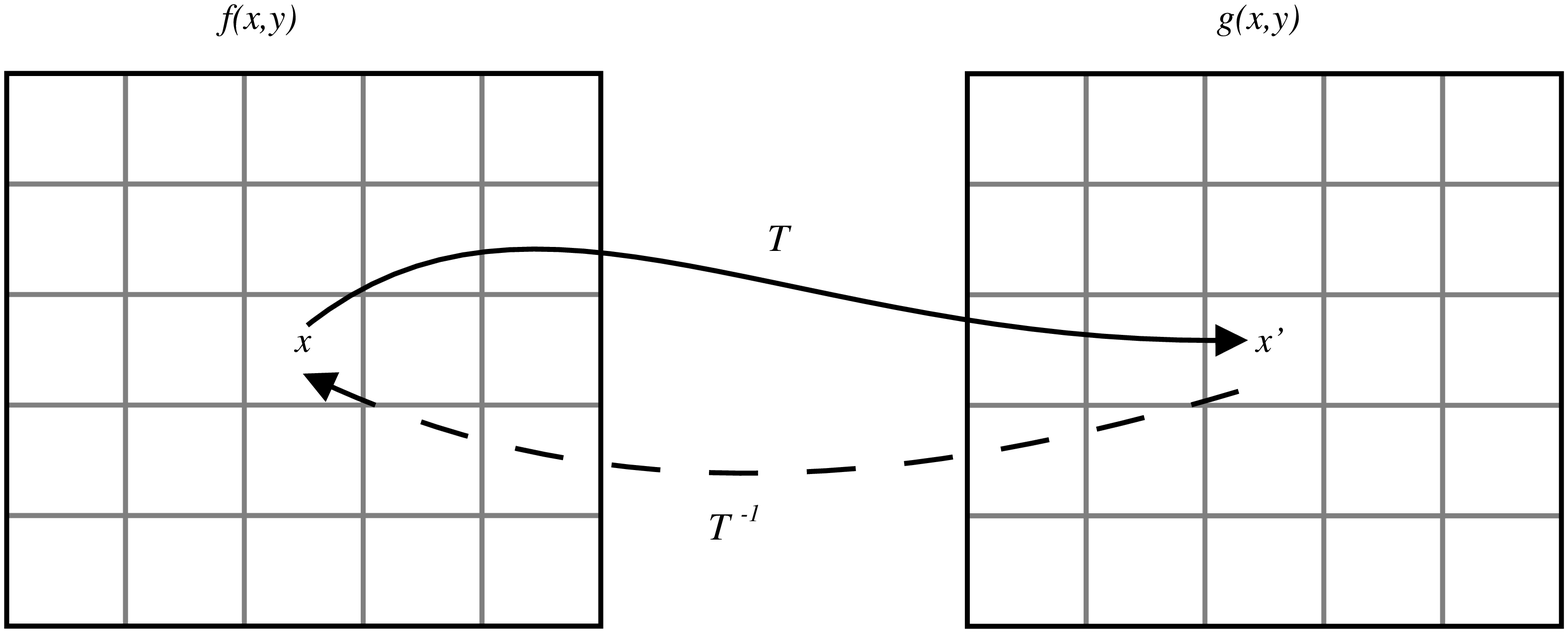}
	\caption{Spatial one-to-one transformation}
	\label{fig:fig}
\end{figure}

\begin{figure}[h]
	\centering
		\includegraphics[width=0.45\textwidth]{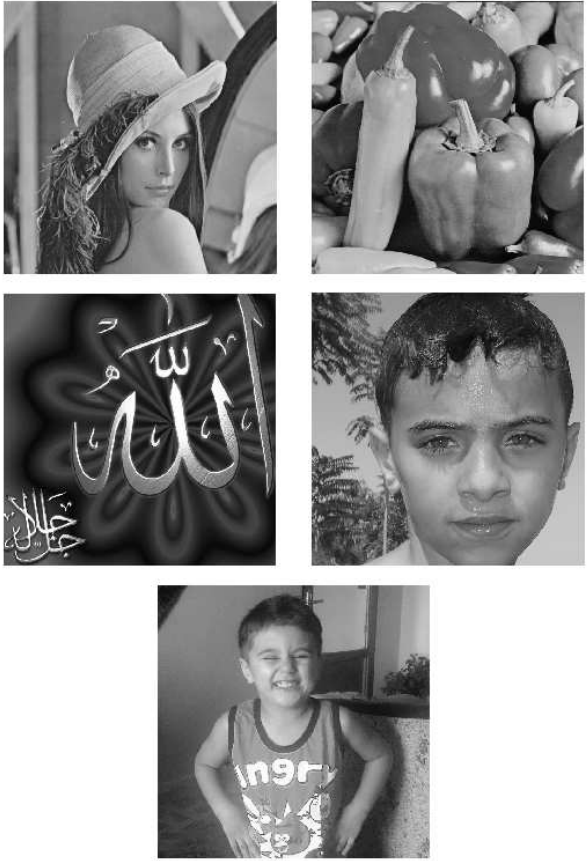}
	\caption{Test Images (Lena,Peppers, Allah, Boy1, and Boy2)}
	\label{fig:fig}
\end{figure}

\section{Proposed Transformation}
\begin{figure*}[!t]
	\centering
		\includegraphics[width=0.90\textwidth]{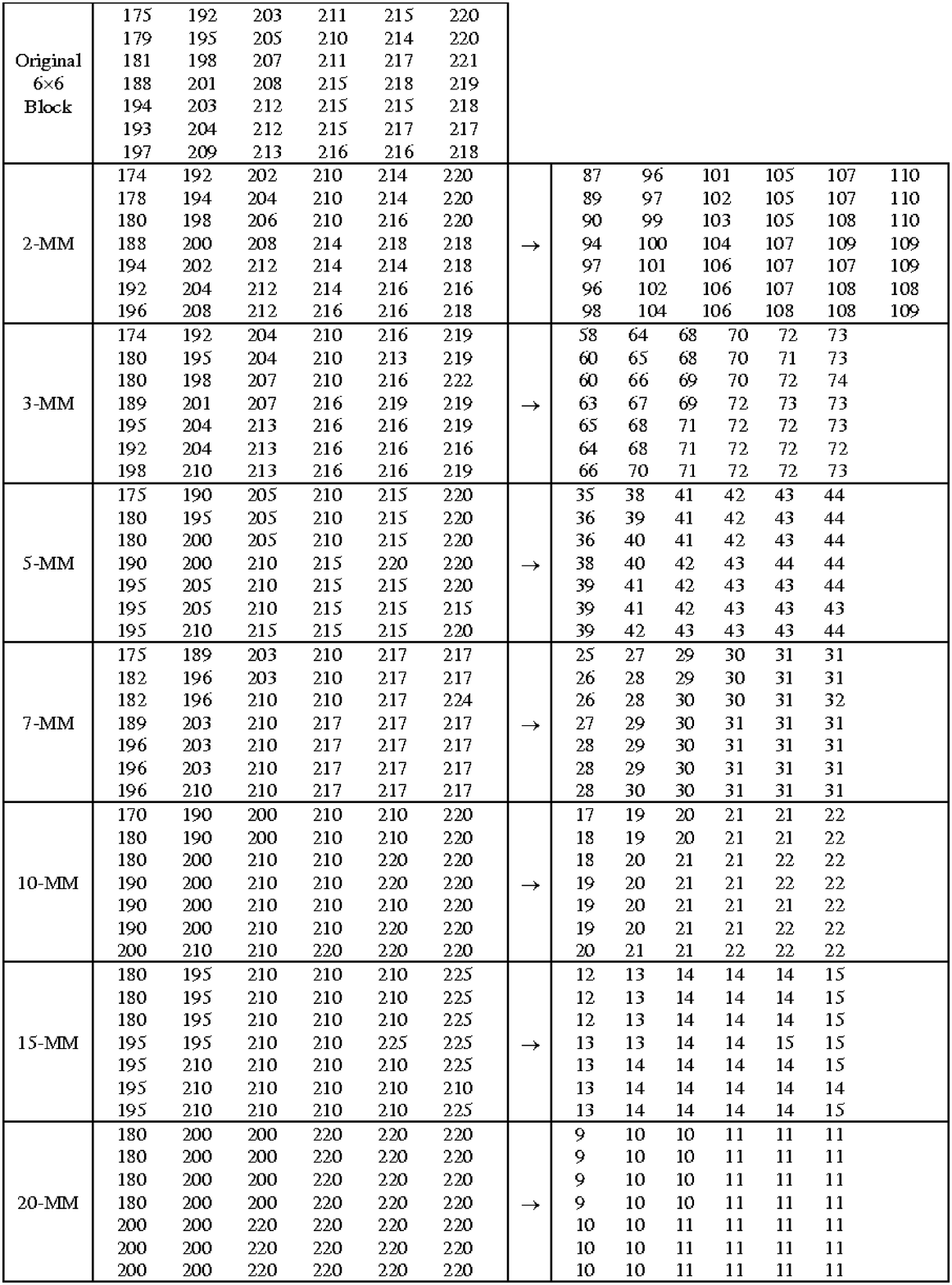}
	\caption{A random 6$\times$6 block from Lena image (Left) $k$-Modulus Method ($k=2,3,5,7,10,15,$ and $20$) (Right) The division of block by the corresponding $k$}
	\label{fig:all}
\end{figure*}

The proposed transform is actually a generalization of the Five Modulus Method (FMM) proposed by \cite{jassim2012five}. The basic idea behind FMM is to transform the original image into multiples of five only. Hence, it was called Five Modulus Method (FMM). Now, a question may arise and that is why not to generalize FMM into $k$-Modulus Method ($k$MM). In another words, instead of using number five only why not to use any integer according to the desired precision?. Therefore, the proposed transform has been established. 
According to \cite{jassim2012five}, the human eye does not differentiate between the original image and the transformed FMM image. Moreover, for each of the Red, Green, and Blue arrays in the color image are consisting of pixel values varying from 0 to 255. The cardinal impression of $k$-Modulus Method is to transform the all pixels within the whole image into multiples of $k$.
According to figure \ref{fig:2to20}, it is clear that the $k$-Modulus Method transformation, up to the 10-Modulus Method, does not affect the Human Visual System (HVS).
Now, to simplify the above idea, an illustrative example will be presented. Suppose $k$=2, then the original range of values 0 to 255 will be transformed as 0, 2, 4, 6,\ldots , 254. Similarly, when $k$=3, the transformed range is 0, 3, 6, 9, 12, \ldots , 255. The same previous procedure may be applied for any integer $k$. The $k$-Modulus method have been applied to $k$=2,3,4,\ldots 20 have been calculated and presented in figure \ref{fig:all}. Obviously, increasing $k$ will lead to high distortion in the output image. Hence, by using try and error, we can see that up to $k$=10 the human eye could not differentiate between the original and the transformed images. According to figure \ref{fig:all}, it can be seen that a selective $k$ values (2, 3, 5, 7, 10, 15, 20) have been used to transform a random 6$\times$6 block from Lena image using $k$-Modulus Method.

\subsection{$k$-Modulus Method Bit Depth}
Bit depth refers to the color information stored in an image. More colors could be stored in an image whenever there is high bit depth in the image. In the case of black and white, it contains one bit either 0 or 1. Hence, it can only show two colors which are black and white. Furthermore, an 8 bit image can store $2^{8}$ which is equal to 256 possible colors, etc. It must be mentioned that, the bit depth specify image size. As the bit depth increase, image size also increases because each pixel in the image requires more information \cite{Gonzalez:2001:DIP:559707}. 
One way to think about bit depth is to consider the difference between having the capability to make measurements with a ruler that is accurate to the nearest millimeter, compared with one that can only measure to the nearest centimeter \cite{evening2010adobe}.

In this paper, a general formula to extract the exact bit depth for each of the $k$-Modulus Methods has been derived as:
\begin{equation}
n=\left\lfloor \frac{256}{k} \right\rfloor +1
\end{equation}

where $k$ is an integer number ($k=2,3,$ \ldots ).

\begin{table}[htbp]
	\caption{Bit depth of $k$-Modulus Method}
	\label{tab:BitDepth}
	\centering
		\begin{tabular}{cccc}
		\hline
			$k$-Modulus Method	& Range	& Binary representation	& Length of pixel \\
			\hline
			2-MM	& 0..128	& 10000000	& 8 \\
			3-MM	& 0..85	& 1010101	& 7 \\
			4-MM	& 0..64	& 1000000	& 7 \\
			5-MM	& 0..51	& 110011	& 6 \\
			6-MM	& 0..42	& 101010	& 6 \\
			7-MM	& 0..36	& 100100	& 6 \\
			8-MM	& 0..32	& 100000	& 6 \\
			9-MM	& 0..28	& 11100	& 5 \\
			10-MM	& 0..25	& 11001	& 5 \\
			11-MM	& 0..23	& 10111	& 5 \\
			12-MM	& 0..21	& 10101	& 5 \\
			13-MM	& 0..19	& 10011	& 5 \\
			14-MM	& 0..18	& 10010	& 5 \\
			15-MM	& 0..17	& 10001	& 5 \\
			16-MM	& 0..16	& 10000	& 5 \\
			17-MM	& 0..15	& 1111	& 4 \\
			18-MM	& 0..14	& 1110	& 4 \\
			19-MM	& 0..13	& 1101	& 4 \\
			20-MM	& 0..12	& 1100	& 4 \\
			\hline

		\end{tabular}
\end{table}

As with resolution, bit depth determines file size. The higher the depth, the greater the file size. It must not be confounded with the amount of actual colors within an image. Therefore, an image with 25 colors may be created with 16 colors. Hence, we may have thousands instead of millions of possibilities. This would obviously lead to increase in file size and that may not be necessary. A versed understanding of bit depth is essential to any graphic or multimedia application \cite{burger2009principles}.

\section{Experimental Results}
\begin{figure*}[!t]
	\centering
		\includegraphics[width=0.80\textwidth]{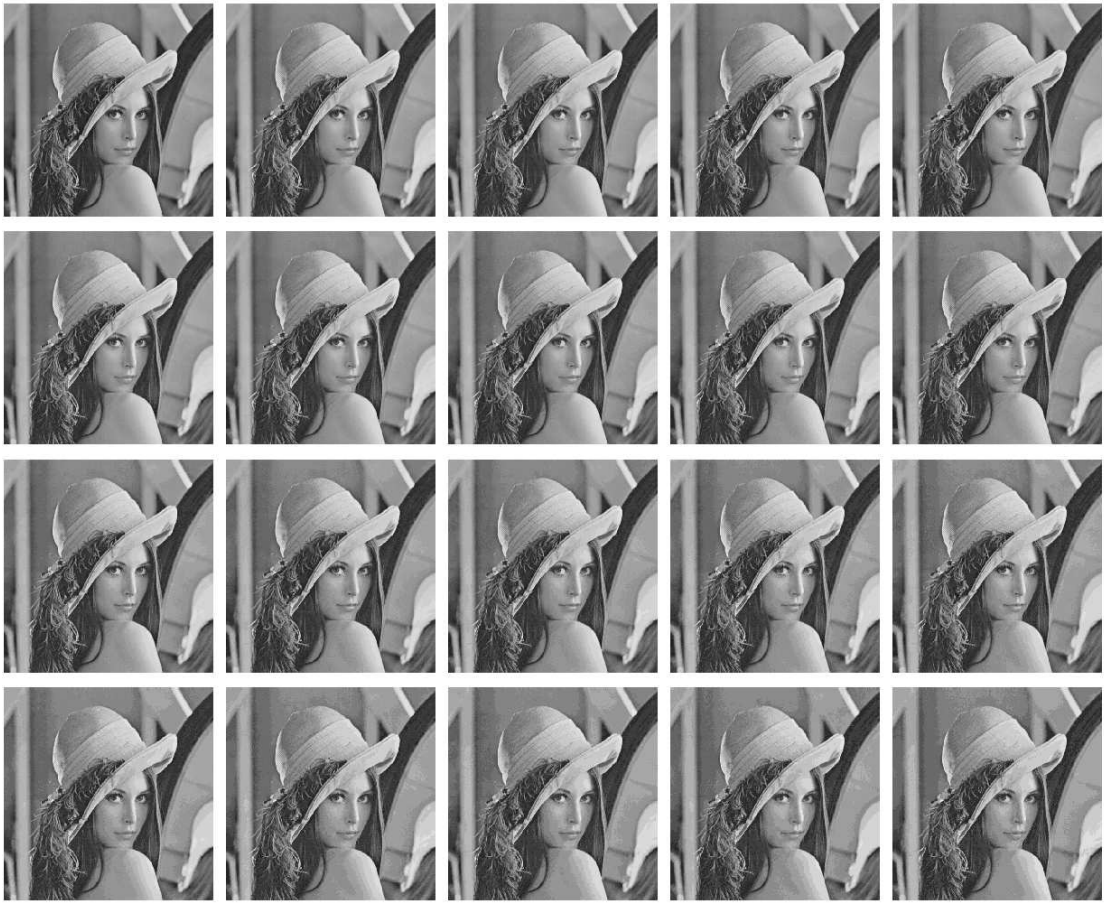}
	\caption{$k$-Modulus Method (First row) Original, 2-MM, 3-MM, 4-MM, 5-MM (Second row) 6-MM, 7-MM, 8-MM, 9-MM, 10-MM
	(Third row) 11-MM, 12-MM, 13-MM, 14-MM, 15-MM (Fourth row) 16-MM, 17-MM, 18-MM, 19-MM, 20-MM  }
	\label{fig:2to20}
\end{figure*}

\begin{figure}[h]
	\centering
		\includegraphics[width=0.45\textwidth]{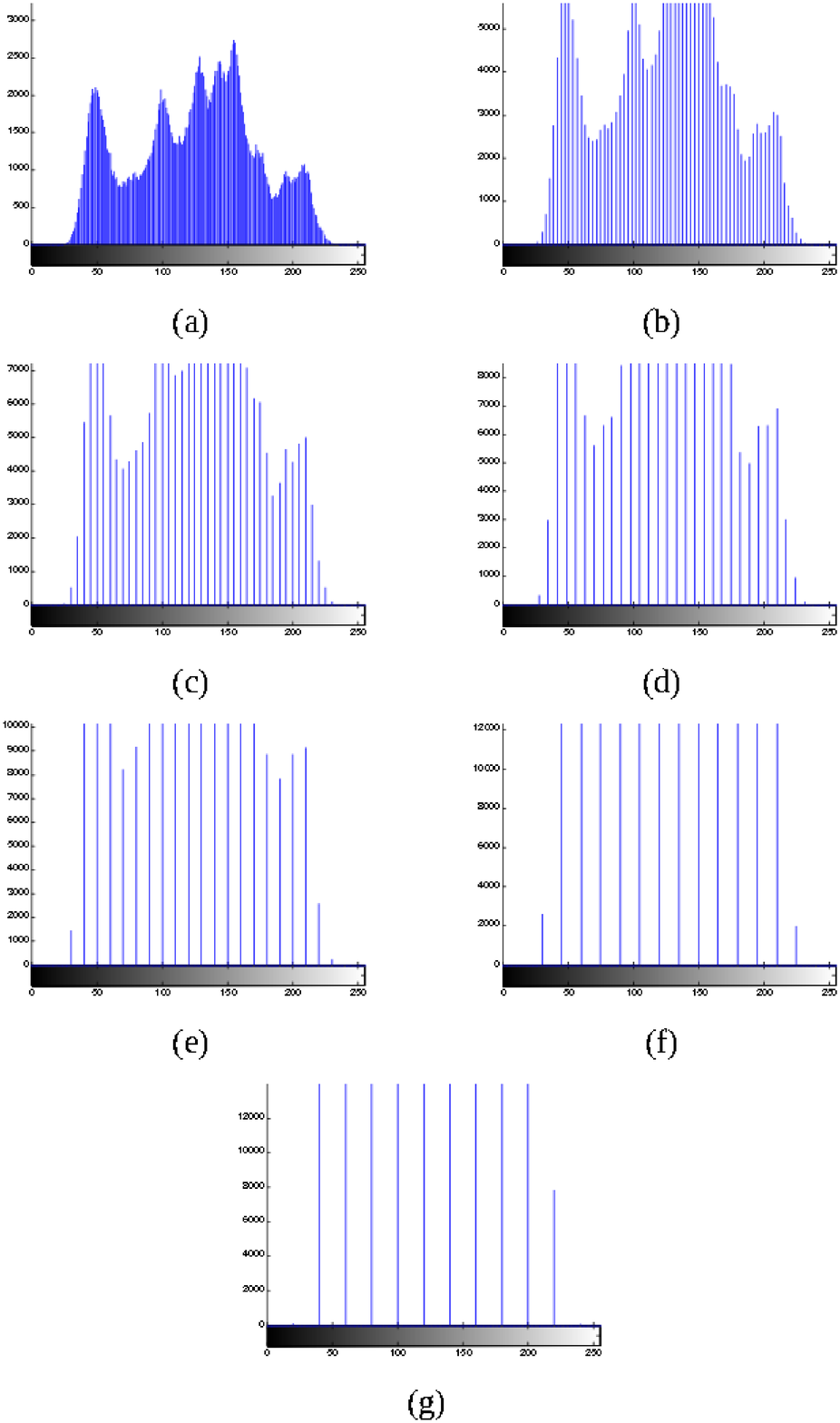}
	\caption{Image Histogram (a) Original (b) 3-MM (c) 5-MM (d) 7-MM (e) 10-MM (f) 15-MM (g) 20-MM}
	\label{fig:bars}
\end{figure}

\begin{figure}[h]
	\centering
		\includegraphics[width=0.34\textwidth]{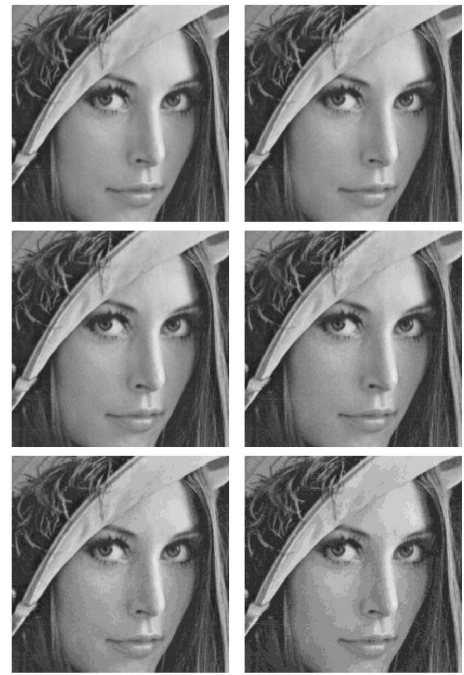}
	\caption{Top left (3-MM) Top right (5-MM) Middle left (7-MM) Middle right (10-MM) Bottom left (15-MM) Bottom right (20-MM)}
	\label{fig:lena1}
\end{figure}

In this article, five test images have been tested using the proposed $k$-Modulus Method. In figure \ref{fig:2to20}, the implementation of the $k$-Modulus Method was applied using $k$=2,3,…,20 for Lena image. It is clear that, the human eye could distinguish dissimilarities up to $k$=10. Higher $k$ may produce noticeable distortion. Moreover, for more illustration, a magnified image of Lena for $k$=3,5,7,10,15,20 have been presented in figure \ref{fig:lena1}. Also, according to figure \ref{fig:bars}, the image histogram for the same previous $k$ values $k$=3,5,7,10,15,20 have been obtained. Here, the same assumption that was stated previously, and that is, up to $k$=10 the shape of the histogram is approximately similar to the original histogram. Finally, The Peak Signal to Noise Ratio (PSNR) \cite{citeulike:2659915} was used to measure the dissimilarities between the transformed and the original images, table (\ref{tab:PSNRKMM}).

\begin{table}[h]
	\caption{PSNR values of $k$-Modulus Method}
	\label{tab:PSNRKMM}
	\centering
		\begin{tabular}{|l|ccccc|}
		\hline
		k-MM & Lena	& Peppers	& Allah	& Boy1	& Boy2 \\
		\hline
		2-MM	& 50.7787	& 50.4276	& 51.1058	& 50.9156	& 51.1518 \\
		3-MM	& 49.5941	&49.1845	& 49.8984	& 49.6841	 & 49.8853 \\
		4-MM	& 46.0309	& 45.7001	& 46.4233	& 46.1599	& 46.3588\\
		5-MM	& 44.7639	& 44.4059	& 45.0469	& 44.9447	& 45.1021 \\
		6-MM	& 42.7984	& 42.4311	& 43.0906	& 42.9874	& 43.1331 \\
		7-MM	& 41.7680	& 41.4927	& 41.8433	& 41.9210	& 42.0895\\
		8-MM	& 40.4774	& 40.0450	& 40.6093	& 40.6246	& 40.7477\\
		9-MM	& 39.5383 &	39.2124	& 39.7726	& 39.9181	& 39.9031\\
		10-MM	& 38.4859	& 38.1514	& 38.8397	& 38.7334	& 38.8230\\
		11-MM	& 37.8295	& 37.4656	& 38.5854	& 38.2731	& 38.1404\\
		12-MM	& 36.9491	& 36.7879	& 37.2798	& 37.0228	& 37.2974\\
		13-MM	& 36.4654	& 35.8355	& 36.9229	& 36.5131	& 36.6772\\
		14-MM	& 35.9254	& 35.3525	& 35.9268	& 35.9884	& 35.9164\\
		15-MM	& 35.0014	& 34.8275	& 35.1099	& 35.4527	& 35.4560\\
		16-MM	& 34.7193	& 34.4729	& 35.2276	& 34.6802	& 34.8957\\
		17-MM	& 33.9161	& 33.7502	& 34.7368	& 34.3753	& 34.2375\\
		18-MM	& 33.6590	& 32.9851	& 33.8404	& 33.7727	& 33.7870\\
		19-MM	& 33.0598	& 32.7615	& 33.2398	& 32.9910	& 33.4654\\
		20-MM	& 32.5316	& 32.3079	& 33.0107	& 32.8999	& 32.8903\\
		\hline
		
		\end{tabular}
\end{table}

According to \cite{welstead1999fractal}\cite{barni2006document}, the typical values for the PSNR in lossy image and video compression are between 30 and 50 dB, where higher is better . Hence, it is clear that the PSNR values in table \ref{tab:PSNRKMM} are quite acceptable.

\section{Conclusion}
In this paper, a novel spatial image transform has been presented. The proposed transform use modulus operator in a way that transform the whole image array into multiples of one and only one integer $k$. As a main conclusion from this article is that, the $k$-Modulus Method could helps in image compression as a scheme but not as stand alone image compression technique. The graphical examples demonstrated that the $k$-Modulus Method produce better results when $k$ is up to 10. Depending on the application used, the designer may control $k$. Therefore, higher $k$ (more than 10) could be used whenever there is a need to a low resolution images.
%\section*{Acknowledgment}
%The authors would like to thank Mrs. Hind E. Qassim for her moral support and some useful help during preparation this %article.
\bibliographystyle{IEEEtran}
\bibliography{bibfile}
\end{document}